\documentclass[letterpaper]{article} \usepackage{not_aaai2026}  \usepackage{times}  \usepackage{helvet}  \usepackage{courier}  \usepackage[hyphens]{url}  \usepackage{graphicx}    \usepackage{natbib}  \usepackage{caption}     \usepackage{algorithm}
\usepackage{algpseudocode}
\usepackage{adjustbox}
\usepackage{newfloat}
\usepackage{listings}
\DeclareCaptionStyle{ruled}{labelfont=normalfont,labelsep=colon,strut=off} 
\floatstyle{ruled}
\newfloat{listing}{tb}{lst}{}
\floatname{listing}{Listing}
\usepackage[hyperfootnotes=false]{hyperref}
\hypersetup{
  hidelinks=true,
  breaklinks=true,
  colorlinks=true,
  linkcolor={red!50!black},
  citecolor={blue!50!black},
  urlcolor={blue!80!black},
}

\usepackage[utf8]{inputenc} \usepackage[T1]{fontenc}    \usepackage{url}            \usepackage{booktabs}       \usepackage{nicefrac}       \usepackage{microtype}      \usepackage{xcolor}         \usepackage[inline]{enumitem}       \usepackage{xspace}
\usepackage{placeins}
\usepackage{lipsum}
\usepackage{bm}
\usepackage{float}
\usepackage{graphicx}
\usepackage{multirow}
\usepackage{pifont}
\usepackage{longtable}
\usepackage{subcaption}
\usepackage{amsmath}
\usepackage{amsfonts}
\usepackage{chngcntr}
\usepackage[nameinlink]{cleveref}
\usepackage{mycommands}

\usepackage{xfrac}

\title{\ourtitle}

\nocopyright

\author {
\ourauthors
}
\affiliations {
  \ouraffiliations
}

\makeatletter
\begin{document}

\maketitle

\blfootnote{\\This paper was accepted at the 40th AAAI Conference on Artificial Intelligence.
This version contains the supplementary material.
}
\begin{abstract}
  The use of learned dynamics models, also known as \emph{world models}, can improve the sample efficiency of reinforcement learning. Recent work suggests that the underlying \emph{causal graphs} of such dynamics models are sparsely connected, with each of the future state variables depending only on a small subset of the current state variables, and that learning may therefore benefit from \emph{sparsity priors}. Similarly, \emph{temporal sparsity}, i.e.\ sparsely and abruptly changing local dynamics, has also been proposed as a useful inductive bias. In this work, we critically examine these assumptions by analyzing
ground-truth dynamics from a set of robotic reinforcement learning
environments in the MuJoCo Playground benchmark suite, aiming to
determine whether the proposed notions of state and temporal sparsity actually tend to hold in typical reinforcement learning tasks.
We study
\begin{enumerate*}[label=(\roman*)]
\item whether the causal graphs of environment dynamics are sparse,
\item whether such sparsity is state-dependent, and
\item whether local system
dynamics change sparsely.
\end{enumerate*}
Our results indicate that global sparsity is rare, but instead the tasks show local, \emph{state-dependent} sparsity in their dynamics and this sparsity exhibits distinct structures, appearing in temporally localized clusters (e.g., during contact events) and affecting specific subsets of state dimensions. These findings challenge common sparsity prior assumptions in dynamics learning, emphasizing the need for grounded inductive biases that reflect the state-dependent sparsity structure of real-world dynamics. 
 \end{abstract}

\section{Introduction}
\label{sec:intro}
Reinforcement Learning (RL) promises to enable robots to learn complex, adaptive behaviors autonomously.
 However, it is often sample-inefficient due to the large number of environment interactions required.

 Model-Based Reinforcement Learning (MBRL) addresses this issue by learning models of the environment’s dynamics, also called \emph{world models} \cite{suttonLearningPredictMethods1988a, schmidhuberLearningThinkAlgorithmic2015,haRecurrentWorldModels2018}.
World models allow agents to simulate interactions and plan effectively with fewer real-world samples  \cite{schmidhuberMakingWorldDifferentiable1990,schmidhuberOnlineAlgorithmDynamic1990,schmidhuberReinforcementLearningMarkovian1990,schmidhuberLearningAlgorithmsNetworks1991,hafnerDreamControlLearning2019,hafnerMasteringDiverseControl2025}.
These world models learn \emph{dense dynamics} where future state values are predicted based on the whole set of the current state values leading to learning spurious correlations between the states which leads to poor generalization and accuracy of prediction \cite{wangTaskindependentCausalState2021}.
To address this issue, recent research has proposed incorporating sparsity as an inductive bias into the learning of dynamics models for Model-Based Reinforcement Learning (MBRL), framing them as \emph{causal models} \cite{wangCausalDynamicsLearning2022, wangBuildingMinimalReusable2024, leiSPARTANSparseTransformer2024}.
These causal models are often assumed to be \emph{sparsely connected}, meaning that each future state variable depends only on a limited subset of the current state variables and actions \cite{ wangCausalDynamicsLearning2022,wangBuildingMinimalReusable2024,hwangFinegrainedCausalDynamics2024, langeCausalityHumanNiche2025}.
Additionally, some studies advocate for modeling \emph{temporal sparsity}, where dynamics change abruptly due to discrete latent transitions in the environment \cite{gumbschSparselyChangingLatent2021, jainLearningRobustDynamics2021a, orujluReframingAttentionReinforcement2025}.

 Despite these promising directions, such assumptions are often validated only in controlled or synthetic environments, e.g. manipulation tasks with unmodifiable state variables and unmovable objects \cite{wangCausalDynamicsLearning2022}.
It remains uncertain whether these sparsity assumptions extend from custom problems with known causal graphs to complex robotic environments featuring contacts and dynamic interactions.

In this paper, we explore the validity of these sparsity-based priors using ground-truth dynamics from the \textit{MuJoCo Playground} \cite{zakkaMuJoCoPlayground2025}, a suite of physics-based robotic benchmarks widely used in RL research.
We focus our analysis on the \emph{sparsity} of the transition functions' \emph{Jacobians}, which capture how future states linearly change with respect to current states and actions.
This analysis enables the characterization of sparsity in the underlying causal structure, as the absence of causal influence corresponds to zero entries in the Jacobian.
Specifically, we address four key questions:
\begin{enumerate}[label=\textbf{(Q\arabic*)}]
	\item Do the true dynamics exhibit sparsity in their causal structure? 
	\item Is such sparsity dependent on the current state?
	\item Do the dynamics undergo temporally sparse transitions?
	\item Does naive Multi-Layer Perceptron (MLP) training recover the ground-truth dynamics' sparsity?
\end{enumerate} 

\section{Related Work}
This section reviews prior research most relevant to our work, beginning with general developments in continuous control environments and reinforcement learning, then narrowing our focus to methods concerned with modeling and learning environment dynamics with sparsity priors.

\paragraph{Continuous Control Environments:}

Continuous control tasks, such as those modeled by MuJoCo \citep{todorovMuJoCoPhysicsEngine2012}, DeepMind Control Suite \citep{tassaDeepMindControlSuite2018}, Brax \citep{freemanBraxDifferentiablePhysics2021} and MuJoCo Playground \cite{zakkaMuJoCoPlayground2025}, have become standard benchmarks for evaluating reinforcement learning (RL) algorithms due to challenging high-dimensional, continuous state and action spaces, requiring agents to learn smooth control policies for locomotion or manipulation.
In this paper, we focus on the continuous control tasks shown in \Cref{fig:envs} from Mujoco Playground.

\paragraph{Continuous Control through RL:}
Model-free approaches, such as DDPG \citep{lillicrapContinuousControlDeep2016}, PPO \citep{schulmanProximalPolicyOptimization2017}, and SAC \citep{haarnojaSoftActorCriticOffPolicy2018}, have achieved impressive results in continuous control.
These methods typically rely on large amounts of interaction data, which limits their sample efficiency.
In contrast, model-based RL approaches seek to improve sample efficiency by learning a predictive model of the environment's dynamics.
Notable methods include PETS \citep{chuaDeepReinforcementLearning2018}, PlaNet \citep{hafnerLearningLatentDynamics2019}, and Dreamer \citep{hafnerDreamControlLearning2019, hafnerMasteringAtariDiscrete2021, hafnerMasteringDiverseControl2025}, which learn latent dynamics models to plan actions or improve policy learning.
Efforts have also explored hybrid model-based/model-free strategies, where learned models are used to generate synthetic rollouts or augment training data. \citep{jannerWhenTrustYour2019}.

\paragraph{World Models and Sparse Dynamics Learning:} 

A growing body of work \citep{pitisCounterfactualDataAugmentation2020, wangTaskindependentCausalState2021,wangCausalDynamicsLearning2022, wangBuildingMinimalReusable2024, hwangFinegrainedCausalDynamics2024, zhaoCuriousCausalitySeekingAgents2025} focuses on learning structured models of the world by combining world model  learning \citep{haRecurrentWorldModels2018} with causal representation learning \citep{scholkopfCausalRepresentationLearning2021}.
The main motivation for these methods is that incorporating causal structure into learning dynamics models can increase generalizability, particularly under distribution shifts or changes in controllable factors \citep{wangTaskindependentCausalState2021}.
Recent papers \citep{wangTaskindependentCausalState2021, wangCausalDynamicsLearning2022, wangBuildingMinimalReusable2024} propose inducing causal structure by incorporating sparsity into dynamics model learning.
These papers argue that dense models, which use all current states and actions to predict future states, are prone to capturing spurious correlations between unrelated features which reduces prediction accuracy and hinders generalization. They propose learning a global context \textit{independent} causal graph and assume that the local dependencies do not change over time.
On the other hand, some further methods have been proposed to learn the fine-grained local \textit{context-specific independence} or context (state) dependent sparsity in the causal graphs by learning local causal models \cite{pitisCounterfactualDataAugmentation2020,pitisMoCoDAModelbasedCounterfactual2022,chitnisCAMPsLearningContextSpecific2021,hwangFinegrainedCausalDynamics2024}.
These papers learn the local causal graphs for different purposes, such as data augmentation for RL agents \cite{pitisCounterfactualDataAugmentation2020,pitisMoCoDAModelbasedCounterfactual2022} or exploration \cite{wangELDENExplorationLocal2023}.
These local causal models are learnt in various ways such as using attention scores \citep{pitisCounterfactualDataAugmentation2020}, examining the Jacobians of the learned dynamics model \citep{wangELDENExplorationLocal2023,zhaoCuriousCausalitySeekingAgents2025}, vector quantization of local subgraphs \citep{hwangFinegrainedCausalDynamics2024,zhaoCuriousCausalitySeekingAgents2025}, or using hard-attention in a Transformer world model with sparsity regularisation \citep{leiSPARTANSparseTransformer2024}.
Similarly, \citet{orujluReframingAttentionReinforcement2025} use RL agents to dynamically construct sparse, time-varying causal graphs, instead of the soft, dense connections typical of Transformers.
\citet{gumbschSparselyChangingLatent2021} and \citet{jainLearningRobustDynamics2021a} model sparsity in the temporal evolution of latent states through the use of L0 regularization and variational sparse gating mechanisms respectively.

All the papers discussed in this section so far learn dynamics models in custom environments predominantly with known causal graphs, under the assumption that sparsity provides a useful inductive bias for MBRL in these environments. However, it was unclear if this assumption actually holds for the true dynamics of \textit{common} continuous-control reinforcement learning environments as well. In this paper, we systematically investigate this assumption.
\begin{figure}[tb]
	\centering
	\includegraphics[width=\linewidth]{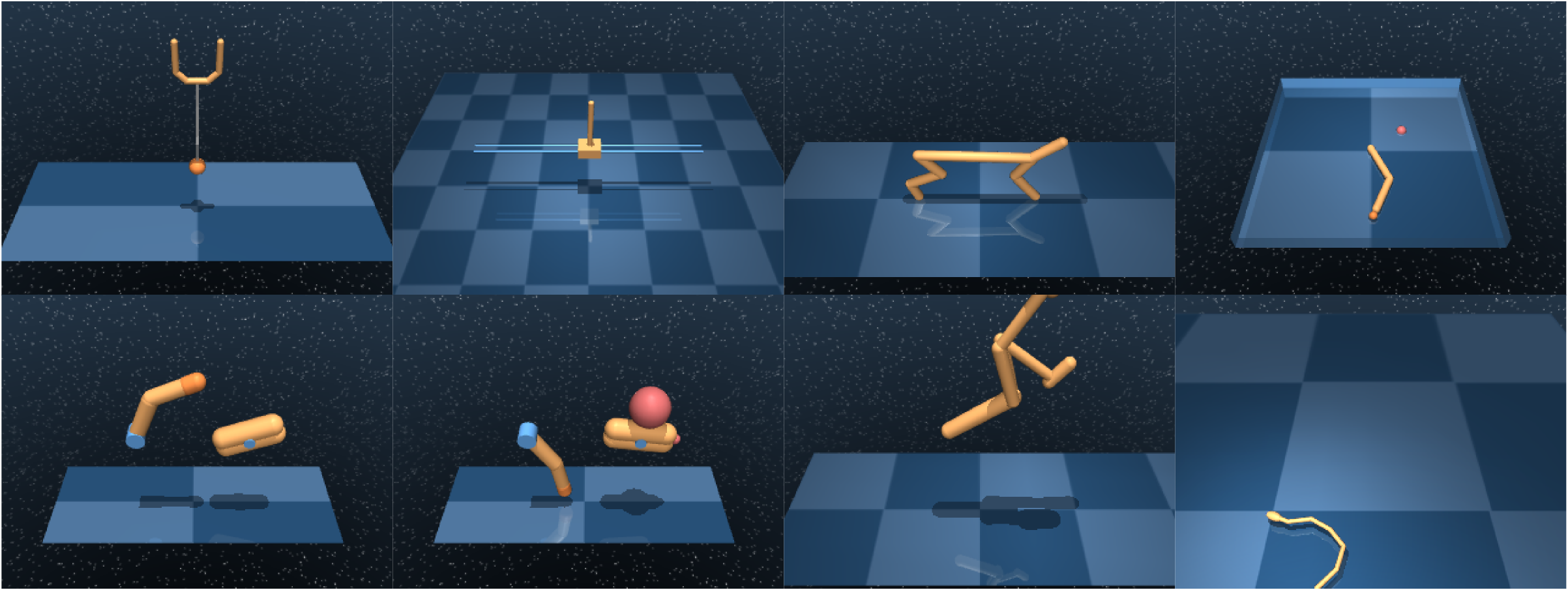}
	\caption{Benchmark Environments from the DeepMind Control Suite \citep{tassaDeepMindControlSuite2018}, implemented in MuJoCo Playground \cite{zakkaMuJoCoPlayground2025}: (top)
		BallInCup, CartpoleBalance, CheetahRun, ReacherHard, (bottom) FingerSpin, FingerTurnEasy, WalkerRun, SwimmerSwimmer6 \  }
	\label{fig:envs}
\end{figure}
 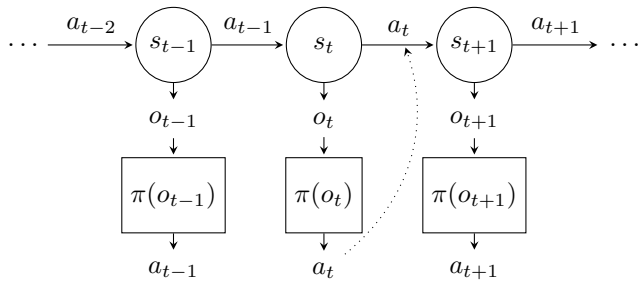
\begin{figure}
  \begin{tikzpicture}[
    ->, >=stealth,
    shorten >=1pt,
    node distance=1cm and 2cm, auto,
    state/.style={circle, draw, minimum size=1cm},
    policy/.style={rectangle, draw, minimum size=1cm}
]

\node (s_t_minus_1) [state] {$s_{t-1}$};
\node (o_t_minus_1) [below of=s_t_minus_1, node distance=1cm] {$o_{t-1}$};
\node (s_t) [state, right of=s_t_minus_1, node distance=2cm] {$s_t$};
\node (o_t) [below of=s_t, node distance=1cm] {$o_t$};
\node (s_t_plus_1) [state, right of=s_t, node distance=2cm] {$s_{t+1}$};
\node (o_t_plus_1) [below of=s_t_plus_1, node distance=1cm] {$o_{t+1}$};

\node (policy_t_minus_1) [policy, below of=o_t_minus_1, node distance=1cm] {$\pi(o_{t-1})$};
\node (a_t_minus_1) [below of=policy_t_minus_1, node distance=1cm] {$a_{t-1}$};
\node (policy_t) [policy, below of=o_t, node distance=1cm] {$\pi(o_t)$};
\node (a_t) [below of=policy_t, node distance=1cm] {$a_t$};
\node (policy_t_plus_1) [policy, below of=o_t_plus_1, node distance=1cm] {$\pi(o_{t+1})$};
\node (a_t_plus_1) [below of=policy_t_plus_1, node distance=1cm] {$a_{t+1}$};

\path (s_t_minus_1) edge node {$a_{t-1}$} (s_t);
\path (s_t) edge node (transition) {$a_t$} (s_t_plus_1);
\path (s_t_minus_1) edge node {} (o_t_minus_1);
\path (s_t) edge node {} (o_t);
\path (s_t_plus_1) edge node {} (o_t_plus_1);

\node (left_ellipsis) at (-2, 0) {$\ldots$};
\node (right_ellipsis) at (6, 0) {$\ldots$};

\path (left_ellipsis) edge node {$a_{t-2}$} (s_t_minus_1);
\path (s_t_plus_1) edge node {$a_{t+1}$} (right_ellipsis);

\path (o_t_minus_1) edge node {} (policy_t_minus_1);
\path (policy_t_minus_1) edge node {} (a_t_minus_1);
\path (o_t) edge node {} (policy_t);
\path (policy_t) edge node {} (a_t);
\path (o_t_plus_1) edge node {} (policy_t_plus_1);
\path (policy_t_plus_1) edge node {} (a_t_plus_1);

\path (a_t) edge [dotted, bend right=35] node {} (transition);

\end{tikzpicture}
   \caption{The simulator's ground-truth-state  $s_t$  is
    advanced by the simulators $\operatorname{step}(s_t, a_t)$
    function to produce the next state $s_{t+1}$. Each of these states $s_t$ non-invertibly produces a
    corresponding observation $o_t$, that is used by the trained agent
    to choose an action. In this work we analyze the dynamics of the ground truth
    next state $s_{t+1}$ with respect to the current state $s_t$.}
  \label{fig:state_structure}
\end{figure}

\section{Background and Notation}

This paper focuses on robotic RL benchmarks
and in particular DM Control Suite environments \cite{tassaDeepMindControlSuite2018}, as implemented in the MuJoCo Playground
  \citep{zakkaMuJoCoPlaygroundOpensource2025}, for the differentiable MJX simulator a version of MuJoCo \citep{todorovMuJoCoPhysicsEngine2012} written in Jax \cite{bradburyJAXComposableTransformations2018}.
The considered systems are vector-valued, discrete time and time-invariant, and their evolution is fully captured by $s_{t+1} = \operatorname{step}(s_t, a_t)$  where $s_t, s_{t+1} \in \mathbb{R}^{d_s}$ are $d_s$ dimensional vectors representing the ground truth states, $a_t \in \mathbb{R}^{d_a}$ is a $d_a$ dimensional vector representing the control signal (a.k.a. \textsl{action}), and $\operatorname{step}:\mathbb{R}^{d_s}\times \mathbb{R}^{d_a}\to \mathbb{R}^{d_s}$ is the simulator's step function\footnote{Up to the indeterminism by
    modern GPU implementations.} that calculates the next state. 
\Cref{fig:state_structure} illustrates how the states change with the actions. 
The observations(eg. pixel space) depend non-invertibly on the states. Performing a similar analysis on observations requires the ground-truth Jacobians between successive observations, which do not exist generally. Thus, we restrict our analysis to state space only.
To interact with the environment, actions are sampled from the stochastic policy of the trained agent, i.e., $a_t \sim \pi(\cdot | o_t)$.
The observations $o_t$ capture some relevant information about the ground-truth state, but the mapping $o_t = f_o(s_t)$ is, in general, not invertible.
Note that all vectors in this manuscript are column vectors, if not stated otherwise.

  \paragraph{Sparse Causal Graphs}

The dynamics model, i.e., the $\operatorname{step}$ function, can be viewed as a vector of scalar-valued component functions, mapping the vector valued inputs to the scalar output of each separate state dimension $i$, i.e.,  $(s_t, a_t) \mapsto s^{(i)}_{t+1}$.
These scalar-valued output functions may depend only on a subset of the input variables $(s'_t, a'_t)$ where ${s'_t \subseteq s_t=\{s^{(1)}_t, \ldots, s^{(d_s)}_t\}}$. 
 If the output for state variable $s^{j}_{t+1}$ does not depend on $s^{i}_t$, then no connecting edge exists in the causal graph representing the $\operatorname{step}$ function.
\emph{Sparse} causal graphs (i.e., graphs with few edges) are in principle simpler to learn than dense ones, since the graph contains less information and the output can be inferred by using only a subset of the inputs. 
 However, the knowledge of which edges are absent is generally unavailable, making it hard to harness this sparsity to simplify the learned model.
  \paragraph{Differentiable Dynamics and Jacobians}

We consider differentiable $\operatorname{step}$ functions, and we define the first order derivatives (a.k.a. \textsl{Jacobians}) w.r.t. states and actions as
\begin{align*}
   J_s = \frac{\delta}{\delta s_t} \operatorname{step}(s, a)
   = & \begin{bmatrix}
     \frac{\delta \operatorname{step}_1}{\delta s^{(1)}_t} & \ldots & \frac{\delta \operatorname{step}_1}{\delta s^{(d_s)}_t} \\
     \vdots & \ddots & \vdots \\
     \frac{\delta \operatorname{step}_{d_s}}{\delta s^{(1)}_t} & \ldots & \frac{\delta \operatorname{step}_{d_s}}{\delta s^{(d_s)}_t}
   \end{bmatrix}\\
J_a = \frac{\delta}{\delta a_t} \operatorname{step}(s, a)
   = & \begin{bmatrix}
     \frac{\delta \operatorname{step}_1}{\delta a^{(1)}_t} & \ldots & \frac{\delta \operatorname{step}_1}{\delta a^{(d_a)}_t} \\
     \vdots & \ddots & \vdots \\
     \frac{\delta \operatorname{step}_{d_s}}{\delta a^{(1)}_t} & \ldots & \frac{\delta \operatorname{step}_{d_s}}{\delta a^{(d_a)}_t}
   \end{bmatrix}
 \end{align*}
\noindent respectively.
 In other words, the Jacobians above capture the local variability of the future state $s_{t+1}$ given infinitesimal variations of the current state $s_t$ and action $a_t$.
The goal of this paper is to study the system's \textsl{Jacobians}, which give useful information about the local interactions between variables, in order to assess the sparsity of the underlying causal graph.
To this end, we collect and analyze a dataset $\mathcal{D}$ consisting of states, actions, next states and corresponding state and action Jacobians.
To make sure that we cover relevant parts of the state space, we collect the dataset by using an expert policy trained via reinforcement learning (\Cref{sec:a:trained_agents}),
and using MJX, we auto-differentiate $\operatorname{step}(s, a)$
  to obtain the state and action Jacobians $J_s$ and $J_a$.

\section{Using Jacobians to Assess Sparsity}
\label{sec:jacobian_sparsity}
To see how the Jacobians of the environment dynamics relate to sparsity in the causal graph of state variables, consider a differentiable function
\begin{align*}
  f: \mathbb{R}^m &\to \mathbb{R}^n\\
  x= (x_1, \ldots, x_m) &\mapsto f(x) = (f_1(x), \ldots, f_n(x)),
\end{align*}
where $x_i$ denotes the $i$-th input variable and $f_j$ the component function mapping $x$ to the $j$-th output variable $y_j$.
Furthermore, suppose the existence of a directed graph $\mathcal{G} = (V, E)$ with vertex set
\[
  V = \{x_1, \ldots, x_m, y_1, \ldots, y_n\}
\]
encoding direct causal relationships from inputs to outputs.
An edge $(x_i \to y_j) \in E$ indicates that the output $y_j$ directly depends on the input $x_i$.
As a consequence, for each $j \in \{1, \ldots, n\}$, $f_j$ is a function only of those inputs $x_i$ for which $(x_i \to y_j) \in E$.
Hence, if there is no such edge, the partial derivative of $f_j$ with respect to $x_i$ must be zero, i.e.,
\[
  (x_i \not\to y_j) \implies \frac{\partial f_j}{\partial x_i} = 0.
\]
Zero partial derivatives expresses that $f_j$ is invariant to infinitesimal changes in $x_i$ when $x_i$ is not a direct cause of $y_j$ in the causal graph $\mathcal{G}$.

Consequently, the absence of edges between certain inputs and outputs in the causal graph $\mathcal{G}$ directly translates into sparsity in the Jacobian matrix:
\[
  J_f(x) := \left(\frac{\partial f_j}{\partial x_i}(x)\right)_{\substack{1 \leq j \leq n \\ 1 \leq i \leq m}} \in \mathbb{R}^{n \times m}.
\]
More precisely, if an edge $(x_i \to y_j)$ is missing in $\mathcal{G}$, the corresponding element of the Jacobian must be zero.
In turn, the number of zero elements present in $J_f(x)$ provides an upper bound on the sparsity of the causal graph.
Each zero element in the Jacobian is a necessary condition for the absence of direct causal influence, but not a sufficient one, due to the possibility of higher order derivatives.

If we consider not only the derivative at a single point, but across the entire domain of $f$ we can make a stronger statement.
A Jacobian element that is zero everywhere, i.e.,
\[
  \frac{\partial f_j}{\partial x_i}(\mathbf{x}) = 0 \quad \forall \mathbf{x} \in \mathbb{R}^m,
\]
implies that all higher-order derivatives with respect to $x_i$ are zero as well.
A global zero element is thus both necessary and sufficient for the absence of a causal edge ($x_i \rightarrow y_j$).
 \section{Experiments}
\begin{table}[bt]
\centering
	\begin{tabular}{@{}l@{}cccc@{}}
		\toprule
		\textbf{Environment} & \multicolumn{2}{@{}c@{}}{\textbf{Dimension}} & \multicolumn{2}{@{}c@{}}{\makecell{\textbf{Jacobian} \\ \textbf{Zero Elements}}} \\
		\cmidrule(r){2-3} \cmidrule(l){4-5}
		& \textbf{State} & \textbf{Action} & \makecell{\textbf{State} (\%)} & \textbf{Action} \\
		\midrule
		BallInCup       & 8  & 2 & 0 (0.00)    & 0 \\
		CartpoleBalance & 4  & 1 & 0 (0.00)    & 0 \\
		CheetahRun      & 18 & 6 & 17 (5.25)   & 0 \\
		ReacherHard     & 4  & 2 & 3 (18.75)   & 0 \\
		FingerSpin      & 6  & 2 & 0 (0.00)    & 0 \\
		FingerTurnEasy  & 6  & 2 & 0 (0.00)    & 0 \\
		WalkerRun       & 18 & 6 & 17 (5.25)   & 0 \\
		SwimmerSwimmer6 & 16 & 5 & 30 (11.72)  & 0 \\
		\bottomrule
	\end{tabular}
	\caption{This table shows the counts of elements in the state
		and action Jacobian that remain constantly zero across
		multiple rollouts, per environment. As this gives an upper
		bound for the sparsity in the underlying global causal graph,
		it indicates that global sparsity is mostly absent from these
		environments.}
	\label{tab:zero_state_elements}
\end{table}

For our experimental analysis, we collect trajectories using expert reinforcement-learning policies trained with PPO \cite{schulmanProximalPolicyOptimization2017}. To encourage exploration, we inject colored noise following the method of \cite{hollensteinColoredNoisePPO2024}. Additional training details are provided in \Cref{sec:a:trained_agents}. The state and action Jacobians of the ground-truth transition function $\operatorname{step}(s,a)$ are obtained via automatic differentiation. We define the \emph{sparsity value} of the Jacobian matrix as the number of zeros in the Jacobian matrix divided by the total number of elements in the Jacobian matrix.

\subsection{(Q1) Do the True Dynamics Exhibit Sparsity in Their Causal Structure?}
\begin{figure}[bt]
	\centering
	\includegraphics[width=.95\linewidth]{{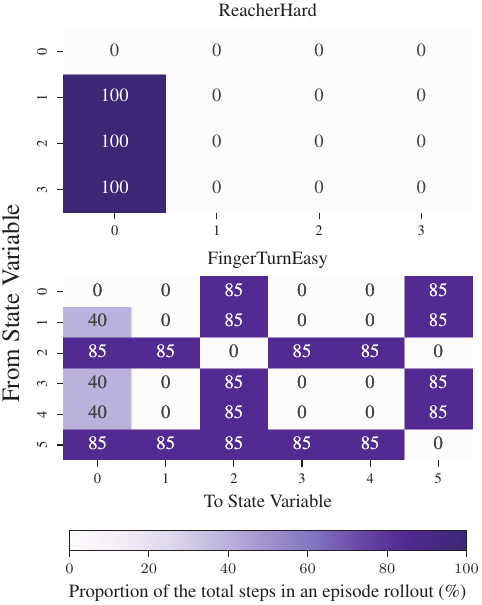}}

\caption{For the two environments ReacherHard and FingerTurnEasy,
		the heatmap illustrates the proportion of time each element of the Jacobians $J_s = \frac{\delta}{\delta s}\operatorname{step}(s, a)$ and $J_a = \frac{\delta}{\delta a}\operatorname{step}(s, a)$ remains \textit{zero} (indicating the independence of the variables) during an episode rollout, expressed as a percentage of the total episode duration averaged across rollouts and seeds. The heatmap values are rounded to the nearest integer.
		Most Jacobian elements remain nonzero throughout the episode, a small number stay zero for the entire duration and the remaining elements are zero for only a fraction of the timesteps. Similar heatmaps for the remaining environments considered for analysis are shown in \Cref{sec:a:sparsity_heatmaps}.  }
	\label{fig:cumulative_sparsity}
\end{figure}
This section investigates whether common Reinforcement Learning environments exhibit globally consistent sparse dynamics.
We examined the Jacobians of $\operatorname{step}$ with respect to
  both $s$ and $a$ for sparsity, i.e., the presence of zero elements that
  persist across all collected samples.
As discussed in \Cref{sec:jacobian_sparsity}, the condition
  \[
    \frac{\partial}{\partial s^{(i)}}\operatorname{step}^{(j)}(s, a) = 0
    \quad \forall\, s, a
  \]
  is both necessary and sufficient for the $(j,i)$ element of
  the state Jacobian to be zero.

Since we cannot exhaustively evaluate this condition, we
  only evaluate a necessary condition for sparsity, and provide an
  upper bound for the true sparsity.
The results for this experiment are listed in \Cref{tab:zero_state_elements}.
We count the zero elements in both the state ($\frac{\delta}{\delta s}$) and action ($\frac{\delta}{\delta a}$) Jacobians.
To account for floating point precision, we use a threshold of $|x|<\sparsityTolerance$ to determine whether an element $x$ is zero.
Our experiments show that there are indeed environments that exhibit globally zero elements in the Jacobians for all samples tested, but that this is only the case for very few elements ($5.25\%-18.75\%$) and only in a handful environments (CheetahRun, ReacherHard, WalkerRun, SwimmerSwimmer6).
This indicates that globally consistent sparsity is rare and is thus unlikely to be a generally important inductive bias.
Instead of requiring a Jacobian element to be zero for all state and action samples, we can investigate the relaxed problem of looking at the percentage of samples where the element is zero.
This is illustrated for the two environments ReacherHard and FingerTurnEasy in \Cref{fig:cumulative_sparsity}.
The results clearly show the globally zero elements in the ReacherHard environment, but further highlight that specific elements can be zero only for a certain percentage of the samples, as for FingerTurnEasy.
This hints at a different, potentially more widely applicable sparsity assumption: state dependent sparsity, which we investigate next.

\subsection{(Q2) Is Such Sparsity Dependent on the Current State?}

\begin{figure}
  \centering
  \includegraphics[width=\linewidth]{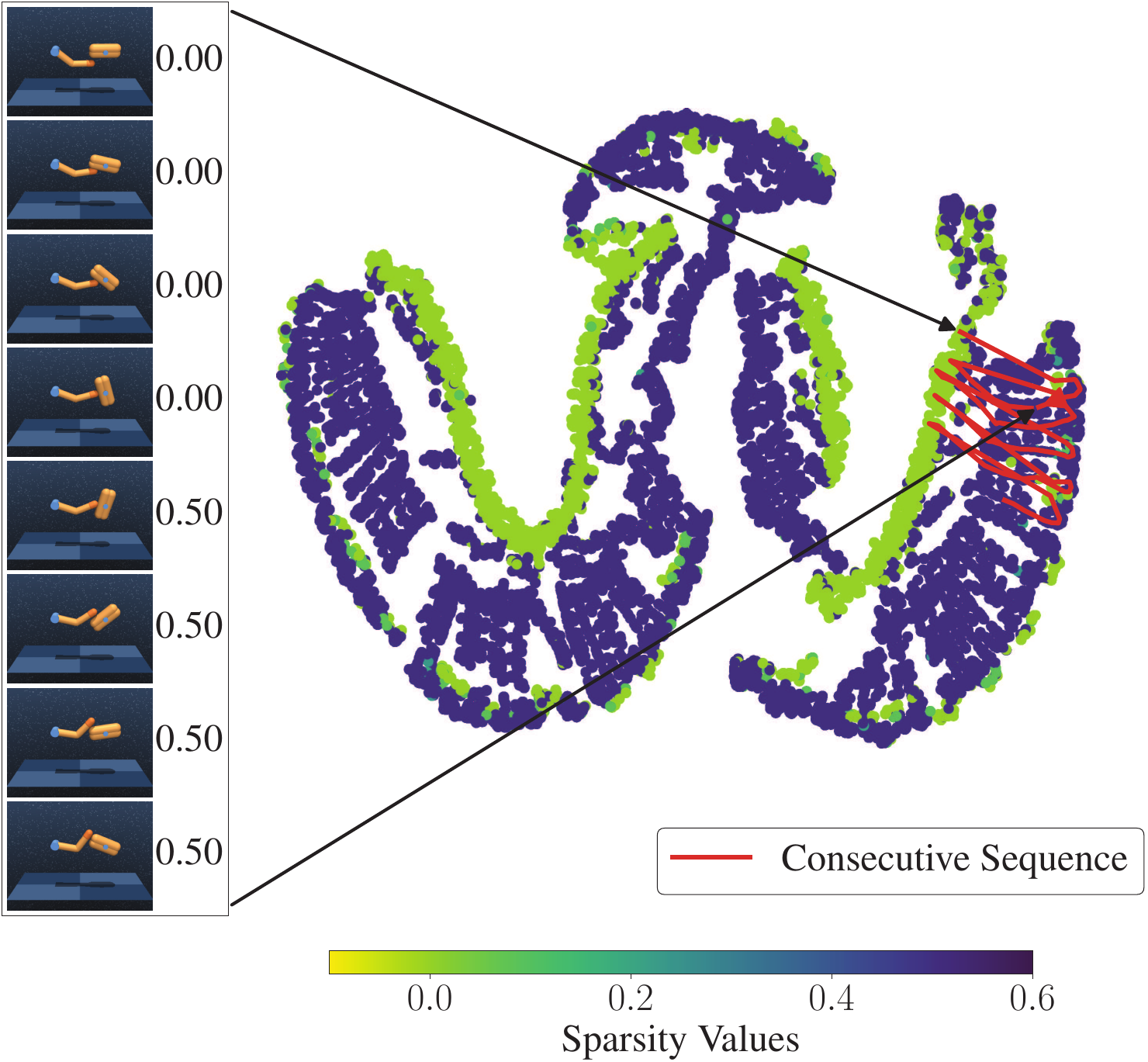}
  \caption{A 2-dimensional t-SNE embedding of state, action and next state tuples with a perplexity value of 50 colored by the \textit{combined} sparsity values of state and action Jacobians across 10 episodes of FingerTurnEasy. Sparsity in the Jacobians is often related to contacts: When the Finger is not moving the object, we observe higher sparsity compared to when the Finger pushes the object in the process. The sparsity values are given near the images of the states. }
  \label{fig:tsne_state_dependent_sparsity_bic}
\end{figure}

\begin{figure}[tb]
  \centering
  \includegraphics[width=\linewidth]{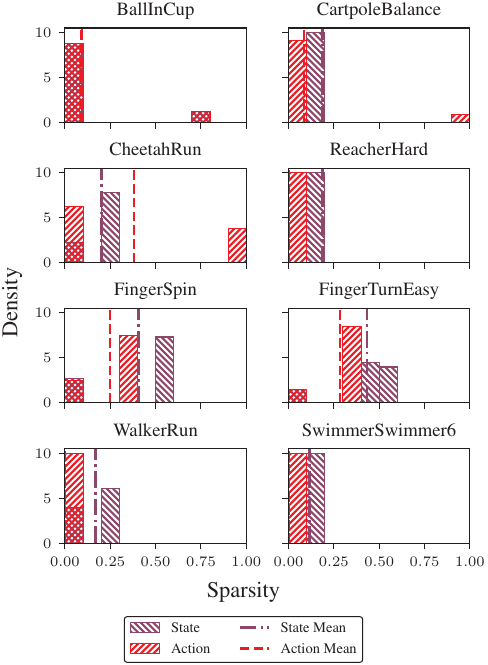}
  \caption{Histograms showing the distribution of  state and action Jacobians sparsity values in the whole dataset with a bin width of 0.1. The sparsity values are mostly concentrated in a small number of bins, indicating a repetition of similar sparsity patterns in the Jacobians over the trajectories. }
  \label{fig:sparsity_histogram_cumulative}
\end{figure}

In the previous section, we examined globally zero elements in
  the Jacobians. We found that some environments do have such zero
  elements, but most do not. Even when they do, there are only a few
  of these globally zero elements.
However, relaxing the requirement of global sparsity for the
  Jacobian elements, we can investigate partial sparsity in the
  Jacobian, i.e., when the Jacobian elements are zero only for
  fraction of the samples.

  Sparsity (i.e., zero elements) in the Jacobian for a fraction of the
  sampled states is a necessary condition for state-dependent
  sparsity, which, as indicated by \Cref{fig:cumulative_sparsity}, is
  present in at least some of the environments.
In this section, we turn to looking at state-dependent sparsity
  in more detail.
\Cref{fig:tsne_state_dependent_sparsity_bic} illustrates state
  dependent sparsity for the FingerSpin environment, through a t-SNE
  embedding of the state, action and next state (s, a, s') tuple  and color coding the combined state and action Jacobian's sparsity at each point.
The figure illustrates a sequence of frames, the corresponding
  transitions through the t-SNE embedded state-space and the sparsity
  values. As can be observed in the first frames, the sparsity is
  lower, when the finger interacts with the object.
This sequence of steps with low sparsity followed by few steps
  of high sparsity, repeats as the object briefly comes into contact with the finger (=low sparsity) and then spins freely (=high sparsity).
Sparsity could be induced in a learned representation, e.g,
  through regularization (similar to \citet{leiSPARTANSparseTransformer2024}. However, this method
  requires tuning the amount of sparsity that is induced.
\Cref{fig:sparsity_histogram_cumulative} illustrates histograms
  of the observed sparsity across all samples for all tested
  environments.
The histograms indicate that the sparsity only assumes certain
  values, presumably based on whether the agent touches the ground
  (CheetahRun, WalkerRun), or interacts with the object (BallInCup,
  FingerSpin, FingerTurnEasy).
The figure also illustrates the mean-sparsity for each
  environment. Presumably, if sparsity regularization is used, it
  should be tuned to induce a similar amount of sparsity on
  average.
While \Cref{fig:sparsity_histogram_cumulative} indicates that
  the sparsity only assumes specific values, it is  unclear how the sparsity changes temporally.

\subsection{(Q3) Do the Dynamics Undergo Temporally Sparse Transitions?}
\begin{figure}[bt]
	\centering
	\includegraphics[width=1.0\linewidth]{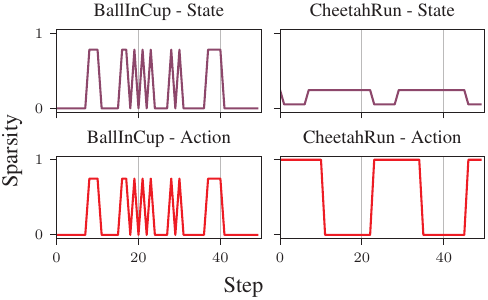}
	\caption{An illustrative example showing how sparsity values evolve over time in the BallInCup and CheetahRun environments.
		In BallInCup, sparsity pattern varies with the tautness of the string connecting the ball and the cup, while in CheetahRun, it reflects changes in the cheetah's gait. Similar plots for the remaining environments considered for analysis are shown in \Cref{sec:a:sparsity_time_evolution}.}
	\label{fig:average_sparsity_evolution}

	\vspace{-15pt} 

\end{figure}

\begin{figure}[bt]
	\centering
	\includegraphics[width=\linewidth]{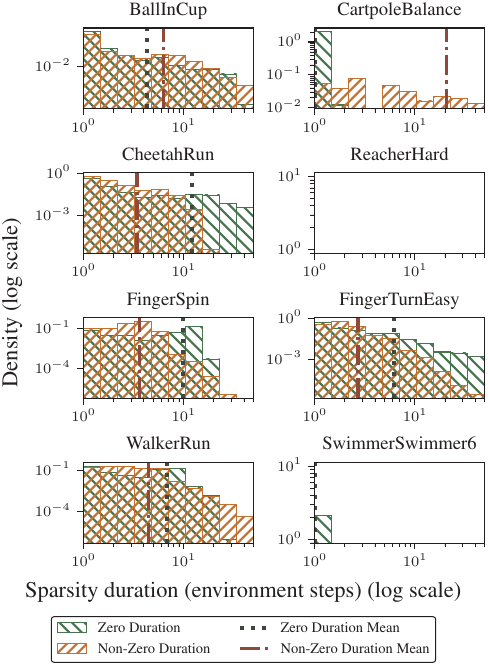}
	\caption{Histograms display the duration distributions of Jacobian elements being zero/non-zero across environments during episode rollouts. The elements mostly remain in the same state over multiple timesteps, indicating temporal sparsity. The Jacobian elements in ReacherHard do not switch between zero/non-zero.
}
	\label{fig:duration_histogram}
\end{figure}
The previous section established that the system assumes
  specific sparsity values, likely due to the structure of the
  interacting objects and the robot, such as objects coming into
  contact or airborne phases of the gait cycle.
If sparsity relates to the gait cycle or contact events, its
  behavior is expected to be temporally consistent---remaining stable
  over multiple steps before switching---which can be observed in \Cref{fig:tsne_state_dependent_sparsity_bic}.
The system's dynamics can be described by differential equations
  where changes in contact correspond to transitions between different
  underlying differential equations. Therefore, incorporating
  inductive biases favoring modelling of temporally sparse switches
  between underlying dynamics may improve the learned performance \cite{gumbschSparselyChangingLatent2021,jainLearningRobustDynamics2021a}.

\citet{gumbschSparselyChangingLatent2021} have pursued a similar
  approach, employing a temporally sparsely changing recurrent network
  to model both agent behavior in partially observable environments
  and environment dynamics. In their method, a hyper-parameter
  penalizes the changes and thus implicitly regulates how often the
  system switches between states, or inversely, stays in the same
  state.
In this section, we examine whether sparsity changes occur in a
  temporally sparse manner and analyze the distribution of durations
  for which these sparsity states persist.
\Cref{fig:average_sparsity_evolution} depicts the evolution of
  the percentage of zero elements over time during a single rollout
  with the expert agent, using the stochastic Gaussian policy. In the
  BallInCup environment, a ball tethered to a controllable cup by a
  string must be caught by moving the cup.
Sparsity is high when the string is loose and low when the ball
  is inside the cup or the string is taut, indicating temporally
  sparse changes.
\Cref{fig:duration_histogram} presents the distribution of durations,
 measured by consecutive timesteps, during which Jacobian elements
 remain in the close to zero or non-zero state. The duration of elements
 which remain zeros for the maximum length of the episode are not shown.
 Many of the elements in these Jacobians remain in the same state for
 durations of multiple steps, hinting at underlying structure that sparsely changes
 in time. The mean duration of change likely poses as a useful default
 hyperparameter for methods attempting to regularize for temporally sparse
 switching.

\subsection{(Q4) Does Naive MLP Training Recover the Ground-truth Dynamics' Sparsity?}

\begin{figure}[bt]
  \centering
  \includegraphics[width=.499\linewidth]{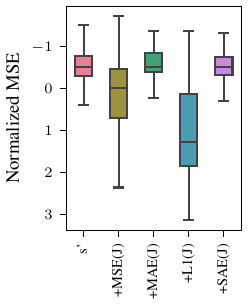}\includegraphics[width=.499\linewidth]{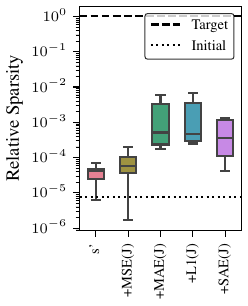}
  \caption{
    Simple MLP architectures are insufficient to capture ground-truth sparsity. The box plots show the quartiles and the median.
    (Left) Test-set next state prediction error aggregated across errors normalized per-environment: adding ground-truth Jacobian loss terms ({\footnotesize $\operatorname{MSE}(J)$, $\operatorname{MAE}(J)$, $\operatorname{SAE}(J)$}) barely affected prediction accuracy (only {\footnotesize $\operatorname{MSE}(J)$} caused a slight reduction), while regularization ({\small $\operatorname{L1}$}) reduced performance. (Right) Test set average MLP Jacobian sparsity value compared to ground-truth. Training increases the MLP Jacobian sparsity value over the untrained ({\small Initial}) value--- even with \emph{ground-truth losses} the sparsity value remains far below the ground-truth ({\small Target}). 
}	\label{fig:q4_relative_sparsity_test_state_loss}
\end{figure}

The previous sections demonstrated that even though global sparsity is rare but sparsity does occur in the ground-truth dynamics, though only to a limited, state-dependent extent.

A natural follow-up question is whether a multi-layer perceptron (MLP) trained to approximate $\operatorname{step}(s, a)$ can capture this sparsity.
To investigate this question, we tested datasets collected from our benchmark environments.
Specifically, we trained a two-layer neural network with ELU activations and 512 units per layer (\Cref{sec:a:learning_experiments}), using mean squared error (MSE) loss to \emph{predict the next state}. Datasets were normalized before training (\Cref{sec:a:data_normalization}).
We consider a Jacobian entry of the MLP with an absolute value below a threshold of $|x|<\networkTolerance$ to be effectively zero.

In addition to the baseline MSE loss, we tested the separate inclusion of ground-truth Jacobian losses (MSE and mean absolute error (MAE)) and an L1 penalty applied exclusively to the predicted Jacobians.
We also introduced a sparsity-aware error (SAE) loss, which applies an MAE loss \emph{only} to the Jacobian elements that are expected to be zero based on the ground-truth Jacobian.
The results, summarized in \Cref{fig:q4_relative_sparsity_test_state_loss}, revealed a slight trend: using an MAE loss for the Jacobian appeared to improve next-state prediction performance, while MSE loss slightly reduced it.
 However, these differences were not significant.
In contrast, the induced sparsity increased significantly when ground-truth Jacobian losses were included, but still remained multiple orders of magnitude below the target sparsity.
Notably, the SAE loss increased sparsity without degrading prediction performance, whereas applying L1 regularization to the Jacobians increased sparsity but negatively impacted state prediction accuracy.
A reduced sparsity value indicates more entangled predictions compared to the ground truth, and consequently higher prediction errors from spuriously learned correlations.
While the naive MLP recovers some sparsity in its predictions, it is insufficient to fully capture the ground-truth sparsity, indicating a need for improved world model architectures.

\section{Discussion}

\paragraph{Main Findings}
Our study provides insights into the role of sparsity in learning the dynamics of classical reinforcement learning (RL) environments and its implications for model-based reinforcement learning (MBRL).
First, we observed that globally sparse causal structures, as indicated by consistently zero Jacobian elements, are rare across most environments:
While some environments, such as ReacherHard and CheetahRun, exhibited limited global sparsity, the majority showed dense interactions between state and action variables.
This suggests that enforcing strong global sparsity priors in learned dynamics models may not be universally beneficial.
Interestingly, we found evidence of state-dependent sparsity i.e. the causal structure of the dynamics changes based on the current state.
For example, in the FingerSpin environment, sparsity was higher when the finger was not in contact with the object and lower during interactions.

In our learning experiments, we observed that while a two-layer MLP trained to approximate $\operatorname{step}(s, a)$ could recover some sparsity in its predictions, the induced sparsity remained far below the target sparsity.
Using MAE-based Jacobian losses slightly improved next-state prediction performance compared to MSE-based losses, though the differences were not statistically significant.
Including ground-truth Jacobian losses increased induced sparsity but did not fully capture the sparsity present in the ground-truth dynamics, leaving the network's predictions more entangled. This likely leads to reduced generalization and more compounding errors during multistep prediction.

\paragraph{Limitations}
While this study provides valuable insights, there are limitations which provide  room for further improvement.
Due to computational budget reasons the experiments were conducted with a limited set of agents and environments. Increasing the set of agents and environments would further generalize the findings.
Similarly, the architectural space is vast and architectural exploration was thus limited to the most intuitive choice, i.e., equal width MLPs.
While expert trajectories are arguably the most relevant data for
modelling the environment while also achieving good task
performance---the collected data were limited to stochastic-policy
rollouts of trained agents---more diverse data could further validate
the results.
\section{Conclusion}
While sparsity priors hold promise for improving sample efficiency and generalization, their effectiveness depends on alignment with the true structure of the environment. Overly strong or misaligned priors could hinder learning by blocking important interactions.
Our findings indicate that sparsity is indeed present in many reinforcement learning environments, but requires modelling in a state dependent way.
Additionally, we observed that changes in sparsity often occur in a temporally sparse manner, with periods of stable sparsity interspersed with abrupt transitions such as contact events or phase changes in locomotion.
Our results also indicate that naive MLP implementations are insufficient to fully capture and exploit these sparsity structures.
These findings highlight the need for further development of dynamics model architectures that can explicitly model sparsity and thus dynamically adapt their causal structure based on state and time, thereby improving generalization and interpretability.

\section*{Acknowledgements}
We thank Christof Beck, Ananth Rachakonda, and Henri Geiß for their
helpful comments on earlier drafts.

This research was partially funded by the Austrian Science Fund (FWF):
I~5755-N~(ELSA), and by
the Autonomous Province of Bolzano-Bozen - South
Tyrol under Funding Agreement 10/2024, Abstractron. \\
\includegraphics[height=\baselineskip,width=!]{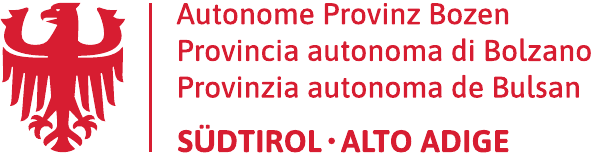}
\includegraphics[height=\baselineskip,width=!]{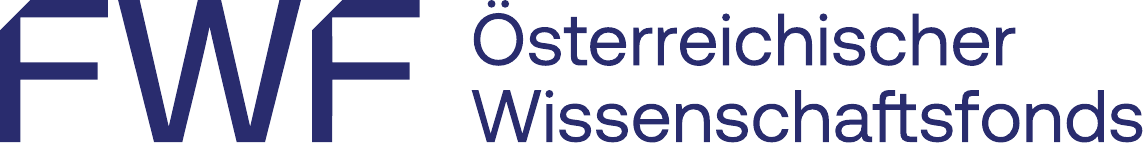}\\

\bibliography{wmwm-processed}

\raggedbottom
\appendix
\section*{Supplementary Appendix}
\addcontentsline{toc}{chapter}{Appendix}

\renewcommand{\thesection}{A\arabic{section}}
\counterwithin{figure}{section}
\counterwithin{table}{section}

\section{Agents \& Data Collection}\label{sec:a:trained_agents}
We generate a dataset of trajectories per environment using pre-trained PPO agents based on the implementation by \cite{raffinStablebaselines3ReliableReinforcement2021}, more specifically following the colored noise methodology proposed by \cite{hollensteinColoredNoisePPO2024}. Multiple agents were trained using the hyperparameter configurations described in \Cref{tab:hyperparameters}. For each environment, the best performing agent was selected. The mean reward achieved across the full episodes in our dataset is shown in \Cref{tab:best_agents}.

Dataset generation then uses the selected best agents. Dataset
generation is necessary because there are no publicly available
datasets containing ground-truth Jacobians of the transition
functions.  For the dataset, we collect a total of $1$ million steps
per environment, using the learned \emph{stochastic} agent policy
(i.e., sampling actions from the parameterized-Gaussian policy). To
improve fidelty MJX and JAX where switched to float64 mode. The MLP
parameters in the learning experiments were represented in float32.

\begin{table}[htbp]
  \begin{tabular}{@{}lrr@{}}
\toprule
Param. & Value & Default \\
\midrule
lr & 5e-05 & 0.0003 \\
n-steps & 64 & 2048 \\
batch-size & 64 &  \\
n-epochs & 10 &  \\
gamma & 0.99 &  \\
gae-lambda & [0.0, 0.95] & 0.95 \\
clip-range & 0.2 &  \\
normalize-advantage & True &  \\
ent-coef & 0 &  \\
vf-coef & 0.5 &  \\
use-sde & False &  \\
sde-sample-freq & -1 &  \\
max-grad-norm & 0.5 &  \\
stats-window-size & 100 &  \\
eval-episodes & 50 &  \\
eval-freq & 10240 &  \\
n-envs & 128 &  \\
total-timesteps & 2.04801e+06 &  \\
seed & [1, 2, 3, 4, 5, 6, 7, 8, 9, 10] &  \\
noise-color & [0.0, 0.5, 1.0, 2.0] &  \\
n-eval-envs & 10 &  \\
\bottomrule
\end{tabular}

   \caption{Best performing agents for each environment were selected among agents trained using these hyperparameter configurations.}
  \label{tab:hyperparameters}

  \begin{tabular}{@{}lc@{}}
    \toprule
    Environment & Mean Reward(Evaluation) \\
    \midrule
    CartpoleBalance & 999.88 \\
    BallInCup & 960.00 \\
    FingerSpin & 902.56 \\
    ReacherHard & 709.20 \\
    SwimmerSwimmer6 & 650.36 \\
    FingerTurnEasy & 515.66 \\
    CheetahRun & 351.75 \\
    WalkerRun & 327.80 \\
    \bottomrule
  \end{tabular}
  \caption{Performance of selected best agents for each environment.}
  \label{tab:best_agents}
\end{table}

\section{Learning Experiments}\label{sec:a:learning_experiments}

\begin{table}[h]
	\centering
  \begin{tabular}{@{}p{0.3\linewidth}p{0.7\linewidth}@{}}
\toprule
 & Experiment Parameters \\
\midrule
activation-function & elu \\
batch-size & 512 \\
dropout-rate & 0.1 \\
env & BallInCup, CartpoleBalance, CheetahRun, FingerSpin, FingerTurnEasy, ReacherHard, SwimmerSwimmer6, WalkerRun \\
epochs & 100 \\
initializer & kaiming\_normal \\
layer-depth & 2 \\
layer-width & 512 \\
learning-rate & 0.002 \\
loss-mode & state, state-jacobian, state-jacobian-l1, state-jacobian-l1-regularizer, state-jacobian-l1-sparse-only \\
normalize-data & True \\
randomize-steps & True \\
seed & 1, 2, 3, 4, 5, 6, 7, 8, 9, 10, 11, 12, 13, 14, 15 \\
test-split & 0.1 \\
weight-decay & 0.001 \\
\bottomrule
\end{tabular}
   \caption{Additional information on the experimental setup.}
  \label{tab:learning-exp-choices}
\end{table}

\subsection{Hyperparameter Search for Dynamics Models}
All hyperparameter search runs use a batch size of $256$, are conducted across $5$ seeds, and otherwise use the same hyperparameters as shown in \Cref{tab:learning-exp-choices}, unless specified otherwise.

\subsubsection{Search Run: 1}
We run a range of experiments across the hyperparemeter selections outlined in \Cref{tab:first_search_range_hyperparams}.
All runs are ranked per environment according to the final state prediction loss on the test set.
For determining our final choice per hyperparmameter, we determine the mean rank across environments and marginalize across all other hyperparameters.
The resulting ranks and choices are shown in \Cref{tab:lr-rank,tab:layer-width-rank,tab:layer-depth-rank,tab:dropout-rate-rank}.

\begin{table}[h]
	\centering
	\begin{tabular}{@{}lc@{}}
		\toprule
		Hyperparameter  & Values  \\ \midrule
		learning-rate   &   [0.01, 0.0001, 0.001] \\
		layer-width     &  [512, 1024]   \\
		layer-depth     &  [4, 6, 2]     \\
		dropout-rate    &  [0.1, 0.5, 0.3]    \\
		\bottomrule
	\end{tabular}
	\caption{Hyperparameter search range for the first run.}
	\label{tab:first_search_range_hyperparams}
\end{table}

\begin{table}[htbp]
	\centering
	\small 

\begin{minipage}[t]{0.45\linewidth}
		\centering
		\begin{tabular}{@{}rr@{}}
\toprule
learning-rate & Test Rank \\
\midrule
\bfseries 0.001 & \bfseries 338.43 \\
0.0001 & 362.32 \\
0.01 & 515.75 \\
\bottomrule
\end{tabular}
 		\captionof{table}{Learning Rate Rank}
		\label{tab:lr-rank}
	\end{minipage}
	\hfill
	\begin{minipage}[t]{0.45\linewidth}
		\centering
		\begin{tabular}{@{}rr@{}}
\toprule
layer-width & Test Rank \\
\midrule
\bfseries 512 & \bfseries 396.23 \\
128 & 399.47 \\
32 & 420.80 \\
\bottomrule
\end{tabular}
 		\captionof{table}{Layer Width Rank}
		\label{tab:layer-width-rank}
	\end{minipage}

	\vspace{1em} 

\begin{minipage}[t]{0.45\linewidth}
		\centering
		\begin{tabular}{@{}rr@{}}
\toprule
layer-depth & Test Rank \\
\midrule
\bfseries 2 & \bfseries 245.32 \\
4 & 437.56 \\
6 & 533.62 \\
\bottomrule
\end{tabular}
 		\captionof{table}{Layer Depth Rank}
		\label{tab:layer-depth-rank}
	\end{minipage}
	\hfill
	\begin{minipage}[t]{0.45\linewidth}
		\centering
		\begin{tabular}{@{}rr@{}}
\toprule
dropout-rate & Test Rank \\
\midrule
\bfseries 0.10 & \bfseries 237.48 \\
0.30 & 415.37 \\
0.50 & 563.64 \\
\bottomrule
\end{tabular}
 		\captionof{table}{Dropout Rate Rank}
		\label{tab:dropout-rate-rank}
	\end{minipage}
	\caption{Achieved ranks in first hyperparameter search.}
\end{table}

\subsubsection{Search Run: 2}
We repeat this procedure with the hyperparameter ranges shown in \Cref{tab:second_search_range_hyperparams}.
All remaining hyperparameters are respectively fixed to the initial values, or to the values found in the first search.
The resulting ranks and choices are shown in \Cref{tab:layer-width-rank-s2,tab:layer-depth-rank-s2,tab:weight-decay-rank-s2}.

\begin{table}[htbp]
	\centering
	\small 

	\centering
	\begin{tabular}{@{}lc@{}}
		\toprule
		Hyperparameter  & Values  \\ \midrule
		layer-width     & [512, 1024]  \\
		layer-depth     & [3, 2]       \\
		weight-decay    & [0.01, 0.001] \\
		\bottomrule
	\end{tabular}
	\captionof{table}{Range of hyperparameters for the second search run.}
	\label{tab:second_search_range_hyperparams}
\end{table}

 \begin{table}[htbp]
 	\centering
 	\small 

\begin{minipage}[t]{0.45\linewidth}
 		\centering
 		\begin{tabular}{@{}rr@{}}
\toprule
layer-width & Test Rank \\
\midrule
\bfseries 512 & \bfseries 19.75 \\
1024 & 21.25 \\
\bottomrule
\end{tabular}
  		\captionof{table}{Layer Width Rank}
 		\label{tab:layer-width-rank-s2}
 	\end{minipage}
 	\hfill
 	\begin{minipage}[t]{0.45\linewidth}
 		\centering
 		\begin{tabular}{@{}rr@{}}
\toprule
layer-depth & Test Rank \\
\midrule
\bfseries 2 & \bfseries 16.61 \\
3 & 24.39 \\
\bottomrule
\end{tabular}
  		\captionof{table}{Layer Depth Rank}
 		\label{tab:layer-depth-rank-s2}
 	\end{minipage}

 	\vspace{1em} 

\begin{minipage}[t]{0.92\linewidth}
		\centering
		\begin{tabular}{@{}rr@{}}
\toprule
weight-decay & Test Rank \\
\midrule
\bfseries 0.00 & \bfseries 20.38 \\
0.01 & 20.62 \\
\bottomrule
\end{tabular}
 		\captionof{table}{Weight Decay Rank}
		\label{tab:weight-decay-rank-s2}
 	\end{minipage}
 	\caption{Achieved ranks in second hyperparameter search.}

 \end{table}
\subsubsection{Search Run: 3}
We perform the same procedure a final time to find a choice of activation function between the Exponential Linear Units (ELU) \cite{clevertFastAccurateDeep2016} and the Rectified Linear Unit (ReLU) \cite{fukushimaVisualFeatureExtraction1969}.
The resulting ranks and choice are shown in \Cref{tab:act-rank-s3}.

\begin{table}[]
	\centering
	\small \centering
	\begin{tabular}{@{}lr@{}}
\toprule
activation-function & Test Rank \\
\midrule
\bfseries elu & \bfseries 8.05 \\
leaky-relu & 12.80 \\
\bottomrule
\end{tabular}
 	\captionof{table}{Activation Function Rank}
	\label{tab:act-rank-s3}
\end{table}

For the learning experiments we double both batch size and learning rate to better utilize our hardware.
We end up with the final hyperparameter choices specified in \Cref{tab:learning-exp-choices}.

\subsection{Data Normalization}\label{sec:a:data_normalization}

We apply Z-score normalization to all variables (states, actions, Jacabions, etc.). For a vector $\mathbf{x} \in \mathbb{R}^N$: 
\[
\tilde{\mathbf{x}} = (\mathbf{x} - \boldsymbol{\mu}_{\text{in}}) \circ \boldsymbol{\alpha}_{\text{in}}, \qquad \boldsymbol{\alpha}_{\text{in}} = \frac{1}{\boldsymbol{\sigma}_{\text{in}} + \varepsilon}
\]
The unnormalized value is recovered as:
\[
\mathbf{x} = \tilde{\mathbf{x}} \circ (\boldsymbol{\sigma}_{\text{in}} + \varepsilon) + \boldsymbol{\mu}_{\text{in}}
\]

For a mapping $f: \mathbb{R}^{N_\text{in}} \to \mathbb{R}^{N_\text{out}}$ (e.g., from actions to next-states), define corresponding mean and standard deviations $(\boldsymbol{\mu}_{\text{in}}, \boldsymbol{\sigma}_{\text{in}})$, $(\boldsymbol{\mu}_{\text{out}}, \boldsymbol{\sigma}_{\text{out}})$. The normalized mapping is:
\[
\tilde{f}(\tilde{\mathbf{x}}) = \left( f(\tilde{\mathbf{x}} \circ (\boldsymbol{\sigma}_\text{in} + \varepsilon) + \boldsymbol{\mu}_\text{in}) - \boldsymbol{\mu}_\text{out} \right) \circ \boldsymbol{\alpha}_{\text{out}}
\]

Let $J = \frac{\partial f}{\partial \mathbf{x}} \in \mathbb{R}^{N_\text{out} \times N_\text{in}}$ (unnormalized Jacobian), and
$\tilde{J} = \frac{\partial \tilde{f}}{\partial \tilde{\mathbf{x}}} \in \mathbb{R}^{N_\text{out} \times N_\text{in}}$ (normalized Jacobian). Then:
\[
\tilde{J} = D_\text{out} \cdot J \cdot D_\text{in}^{-1}
\qquad
D_\text{out} = \mathrm{diag}(\boldsymbol{\sigma}_\text{out} + \varepsilon), \;
D_\text{in} = \mathrm{diag}(\boldsymbol{\sigma}_\text{in} + \varepsilon)
\]
or elementwise,
\[
\tilde{J}_{ij} = J_{ij} \cdot \frac{\sigma^{\text{out}}_i + \varepsilon}{\sigma^{\text{in}}_j + \varepsilon}
\]
with scaling matrix $M_{ij} = \frac{\sigma^\text{out}_i + \varepsilon}{\sigma^\text{in}_j + \varepsilon}$, i.e.,
\[
\tilde{J} = J \circ M
\]

\clearpage
\subsection{Prediction Loss per Environment}
\begin{figure}[h!]
  \includegraphics[width=.99\linewidth]{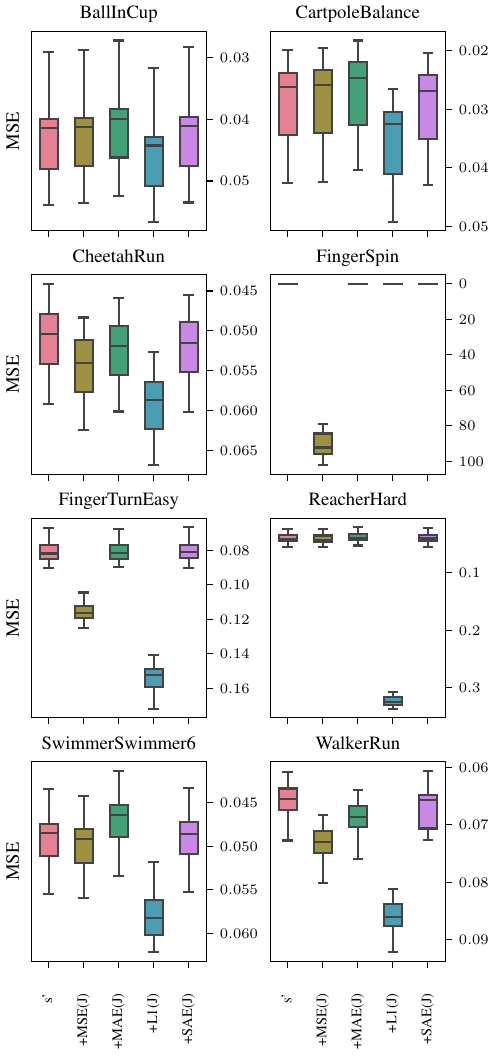}
  \caption{MSE loss of next-state prediction on the test set, of
    different seeds, averaged across training progress.}
\end{figure}
\newpage
\subsection{Sparsity per Environment}
\begin{figure}[h!]
  \includegraphics[width=.99\linewidth]{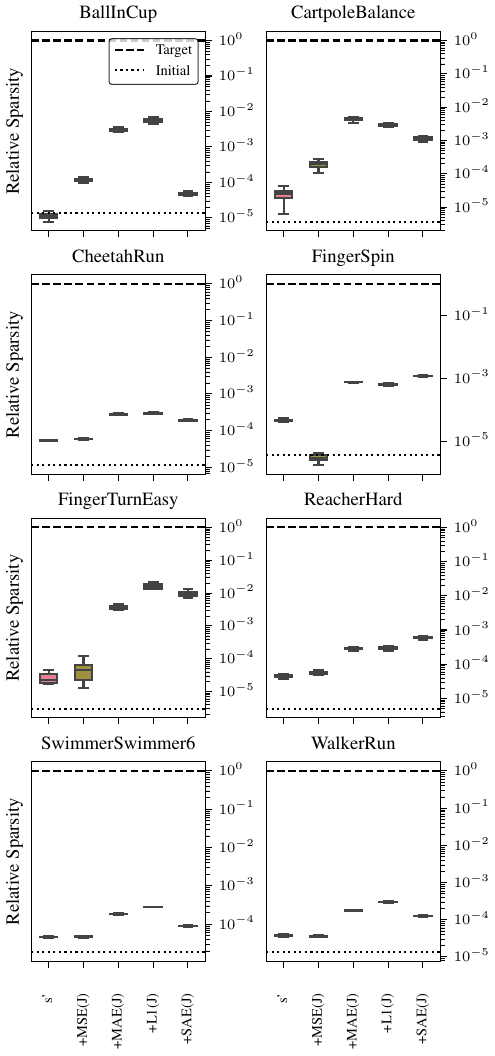}
  \caption{Initial and achieved final sparsity (normalized against the data set target sparsity).}
\end{figure}
\clearpage

\section{Sparsity Plots}
\subsection{Sparsity Heatmaps Showing Zero-Duration Proportions}\label{sec:a:sparsity_heatmaps}
\begin{figure}[h!]
	\centering
	\includegraphics[width=\linewidth]{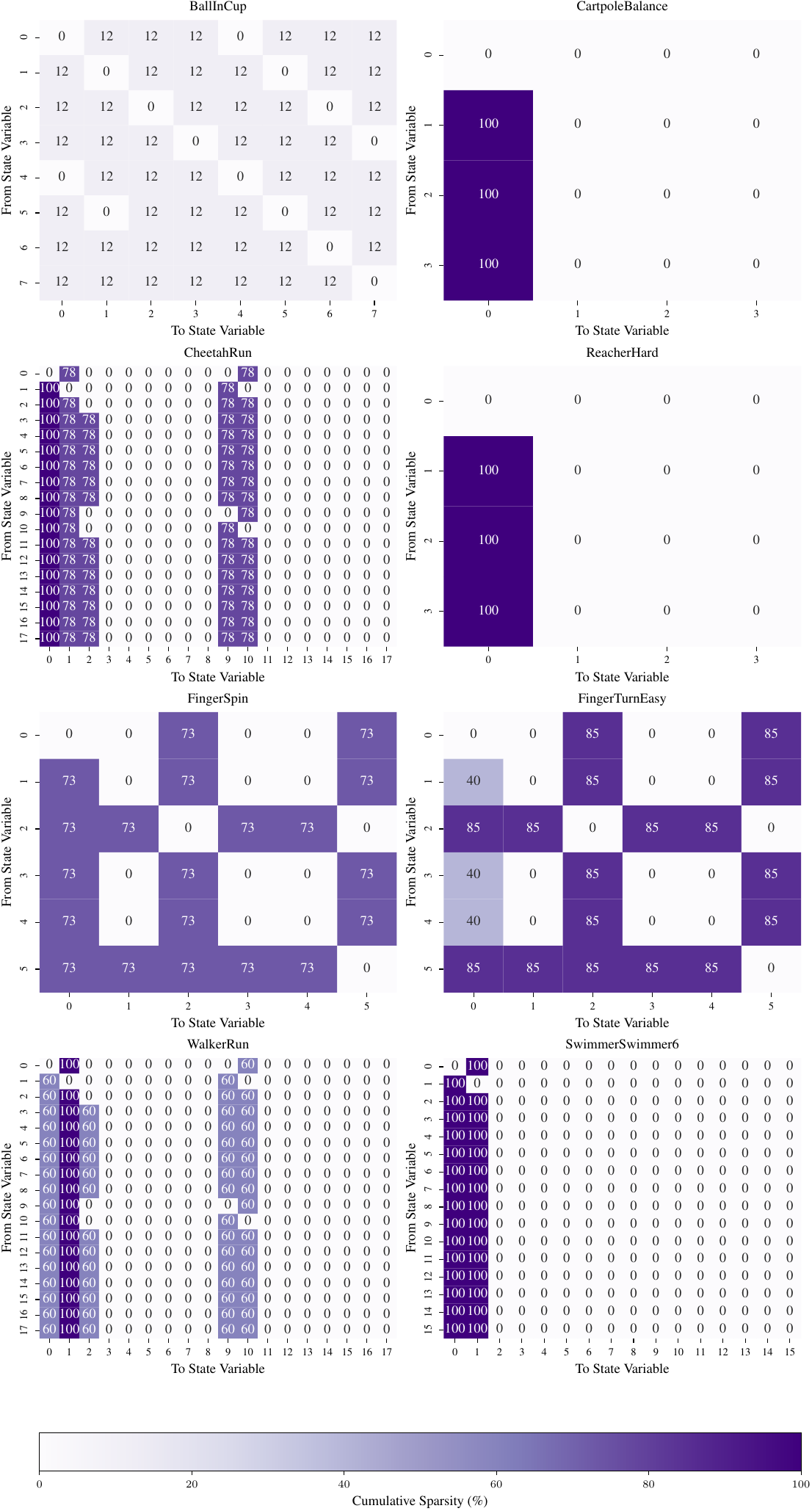}
	\caption{The heatmaps illustrate the proportion of time each element of the state Jacobians $J_s = \frac{\delta}{\delta s}\operatorname{step}(s, a)$  remain \textit{zero} (indicating the independence of the variables) during an episode rollout, expressed as a percentage of the total episode duration averaged across rollouts and seeds. The heatmap values are rounded to the nearest integer.}
	\label{fig:state_sparsity_heatmap_cumulative_all_envs}
\end{figure}

\begin{figure}[h!]
\centering
	\includegraphics[width=\linewidth]{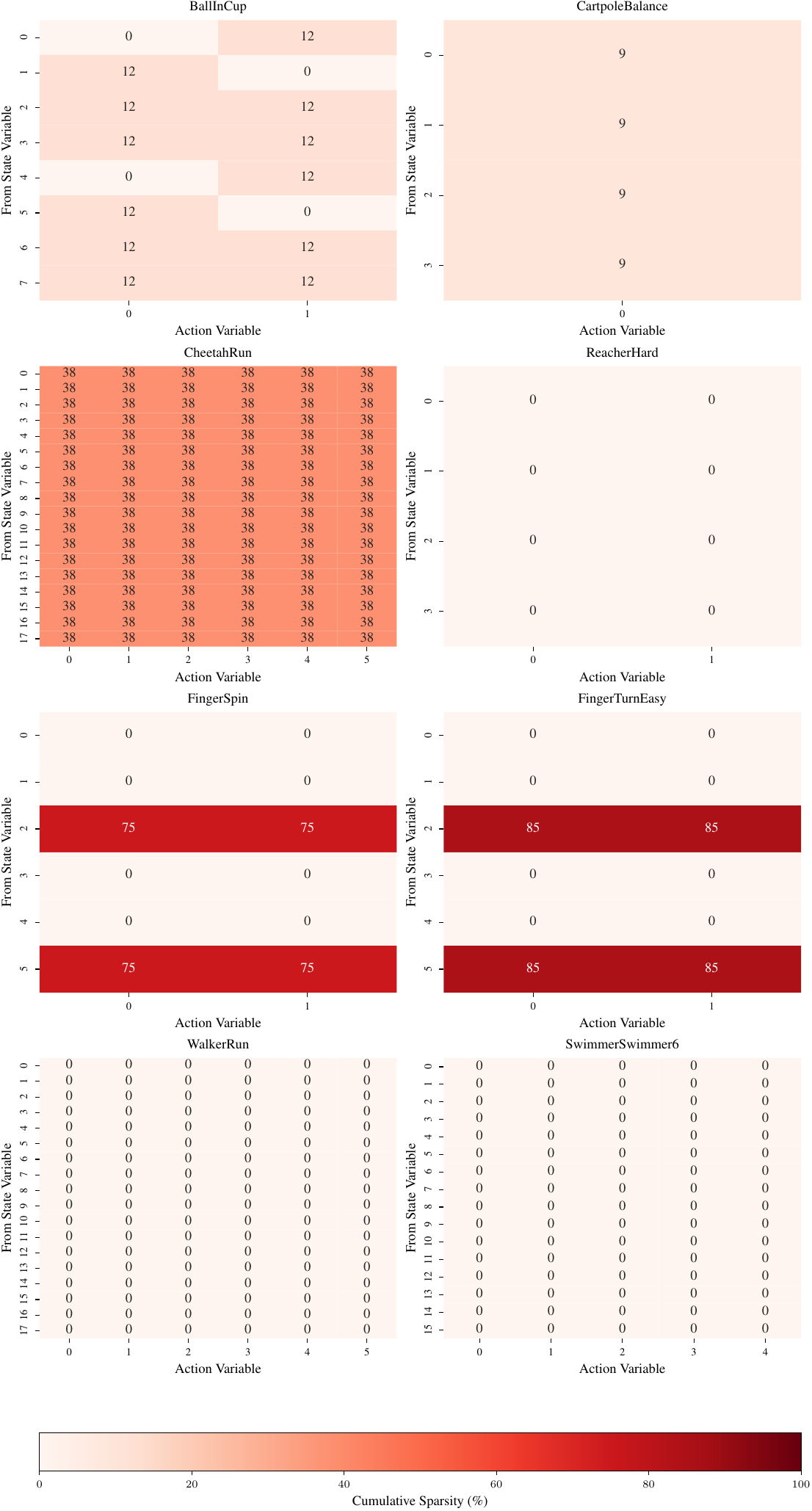}
	\caption{The heatmaps illustrate the proportion of time each element of the action Jacobians  $J_a = \frac{\delta}{\delta a}\operatorname{step}(s, a)$ remains \textit{zero} (indicating the independence of the variables) during an episode rollout, expressed as a percentage of the total episode duration averaged across rollouts and seeds. The heatmap values are rounded to the nearest integer. }
	\label{fig:action_sparsity_heatmap_cumulative_all_envs}
\end{figure}

Sparsity heatmaps showing the duration for which the elements in the state and action Jacobians were zero as a proportion of the total environment steps in a rollout. The following heatmaps are for environments which were not included in the paper for both state and action Jacobians in the \Cref{fig:state_sparsity_heatmap_cumulative_all_envs} and \Cref{fig:action_sparsity_heatmap_cumulative_all_envs}

\clearpage
\subsection{Sparsity time evolution plots}\label{sec:a:sparsity_time_evolution}

\begin{figure}[h!]
	\centering
	\includegraphics[width=\linewidth]{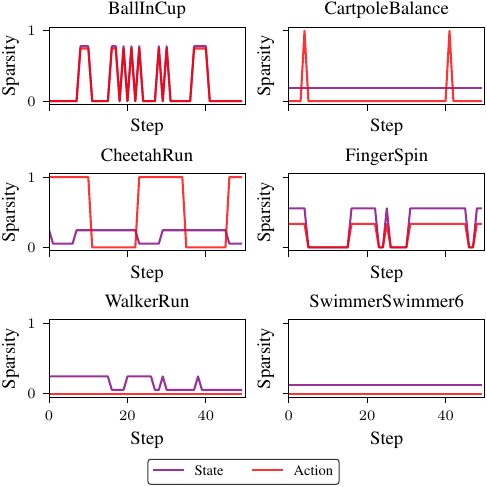}
	\caption{Sparsity values over time for each environment for one single rollout. Only first 50 time steps shown for clarity.}
	\label{fig:sparsity_evolution_all_envs}
\end{figure}

The time evolution of sparsity of state and action Jacobians for the environments not included in the paper are provided here in \Cref{fig:sparsity_evolution_all_envs}.

\subsection{Code and Dataset for training agents}

The code is part of files submitted as part of the supplementary. The code includes the training of the MLPs and the data collection scripts. (partial code released).

Example datasets and their descriptions can be found at this link - https://doi.org/10.5281/zenodo.16743573
 
\end{document}